\documentclass[letterpaper, 10 pt, conference]{ieeeconf}  

\IEEEoverridecommandlockouts                              
\usepackage[utf8]{inputenc}
\usepackage[T1]{fontenc}

\usepackage{multirow}
\usepackage{cite}
\usepackage{tabularx}
\usepackage{adjustbox}

\begin{document}

\title{ Evaluation of the Robustness of Visual SLAM Methods in Different Environments}

\author{Joonas Lõmps\\
University of Tartu, ITS Lab\\
Narva mnt 18, 51009 Tartu, Estonia\\
{\tt\small joonas.lomps@ut.ee}
\and
Artjom Lind\\
University of Tartu, ITS Lab\\
Narva mnt 18, 51009 Tartu, Estonia\\
{\tt\small artjom.lind@ut.ee}
\and Amnir Hadachi\\
University of Tartu, ITS Lab\\
Narva mnt 18, 51009 Tartu, Estonia\\
{\tt\small hadachi@ut.ee}
}

\maketitle

\begin{abstract}
Determining the position and orientation of a sensor vis-à-vis its surrounding, while simultaneously mapping the environment around that sensor or simultaneous localization and mapping is quickly becoming an important advancement in embedded vision with a large number of different possible applications. This paper presents a comprehensive comparison of the latest open-source SLAM algorithms with the main focus being their performance in different environmental surroundings. The chosen algorithms are evaluated on common publicly available datasets and the results reasoned with respect to the datasets' environment. This is the first stage of our main target of testing the methods in off-road scenarios.
\end{abstract}

\section{INTRODUCTION}
Emerging technologies like advances driver-assistance systems (ADAS), autonomous self-driving cars, unmanned and micro-aerial vehicles (UAVs,  MAVs), virtual and augmented reality all share common fundamental building blocks: simultaneous localization and mapping (SLAM) and visual odometry (VO). SLAM techniques try to solve the so-called chicken or egg problem of estimating the robot pose based on a map of an unknown environment while also building the map itself. Visual odometry describes the process of determining the position and orientation of a robot using sequential camera images.

The main sensor of a Visual SLAM (V-SLAM) system is the camera, be it monocular (single camera), stereo (two camera setup) or RGB-D camera (image with depth information per pixel). Cameras are one of the cheapest sensors that provide a rich representation of the environment that enable accurate and robust place recognition. Therefore over the last years research on the possibilities and limitations of V-SLAM has increased. Other sensors are also used to perform or help with SLAM like LiDARs \cite{ren2019robust, lesakodometry, shin2019dvl}, radars \cite{holder2019real, schuster2016landmark} and inertial  measurement  units (IMU) \cite{nutzi2011fusion, leutenegger2013keyframe, vidal2018ultimate}. The higher cost and larger size of the mentioned sensors could be a limiting factor for their use in everyday electronics.

One of the ways to classify V-SLAM is feature-based methods \cite{klein2007parallel}, \cite{ORBSLAM}, that estimate a sparse reconstruction using principal point matching; and direct methods \cite{newcombe2011dtam}, \cite{graber2011online}, \cite{engel2014lsd} which estimate semi-dense or completely dense reconstruction by focusing on high-gradient areas or the direct minimization of the photometric error and TV regularization.

Most of the current state-of-the-art SLAM algorithms are tested in static and feature-rich environments that provide extremely good conditions for the method to perform. Unfortunately, these conditions are rarely seen in real-world use-cases. This can lead to completely different results in non-laboratory conditions where the features are hard to detect, far away or non-existent. As SLAM and VO are complex non-linear estimation problems where minor changes can change the outcome drastically. To get a worthwhile comparison of different algorithms while avoiding manual overfitting for specific environments, the methods should be evaluated in a variety of scenes. Hence, this paper benchmarks the state-of-the-art open-source SLAM systems focusing on their performance in different environmental settings.
\section{VISUAL SLAM METHODS}
In this section, we go over the characteristics and the main logic of the chosen V-SLAM methods. The criteria we were looking at when choosing the V-SLAM methods: designed for monocular or stereo based-camera input, open-source code, released after 2016
\begin{itemize}
  \item Designed for the monocular or stereo camera-based input
  \item Open-source code
  \item Released after 2016
\end{itemize}

The reasons for these limitations are rather straightforward: we are focusing on state-of-the-art V-SLAM methods that we can run in controlled environments. Using non-open-source SLAMs would probably not provide the same results as in the article that introduces them due to the implementation differences, so we decided to only benchmark the methods that provide code themselves.

\subsection{Oriented FAST and rotated BRIEF SLAM}
\begin{figure}[!ht]
\includegraphics[width=\columnwidth]{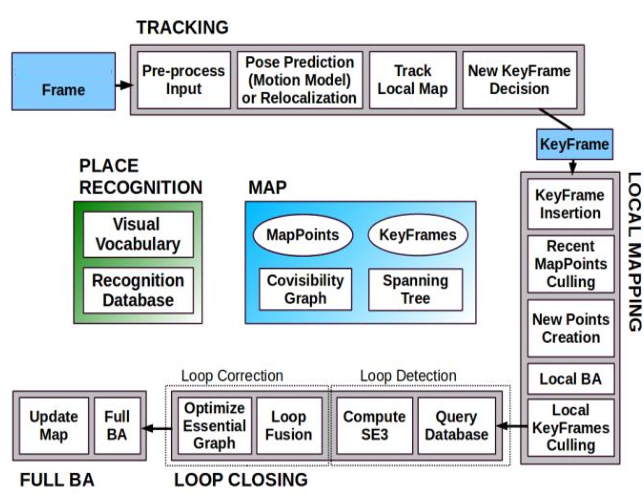}
\caption{Overview of ORB-SLAM2 system, showing all the steps performed by the tracking, local mapping and loop closing threads. Also the main components of the map and place recognition.
\cite{ORBSLAM2}}
\label{ORBSLAM2threads}
\centering
\end{figure}

Oriented FAST and rotated BRIEF SLAM version 2 (ORB-SLAM2) is a feature-based complete SLAM system for monocular, stereo and RGB-D cameras, including map reuse, loop closing and relocalization capabilities that operate in small and large, indoor and outdoor environments \cite{ORBSLAM, ORBSLAM2}. To perform at real-time, binary descriptor ORB \cite{ORBFeatures} is used. ORB descriptor built on top of BRIEF \cite{calonder2010brief} descriptor and FAST \cite{rosten2006machine} keypoint detector, resulting in a rotation invariant and noise resistant descriptor that are extremely fast to compute and match. 
As shown in figure \ref{ORBSLAM2threads} ORB-SLAM2 runs three threads in parallel: tracking, local mapping and loop closing. Tracking localizes the camera with each frame by matching features to the local map and minimizing the reprojection error with motion-only bundle adjustment (BA), local mapping manages the local map and optimized it with local BA, and finally, the loop closing detects large loops and corrects the accumulated drift with pose-graph optimization. A fourth thread is launched to perform full BA after the pose-graph optimization \cite{ORBSLAM2}.

\begin{figure*}[th!]
\includegraphics[width=\textwidth]{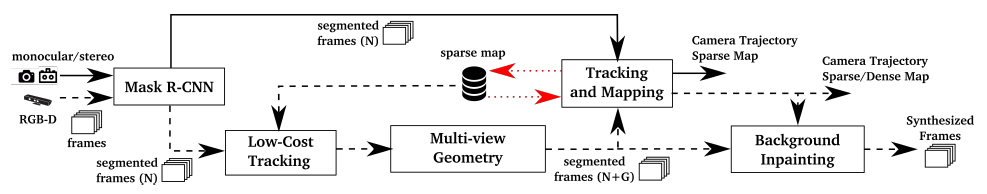}
\caption{ Diagram of the DynaSLAM. Black continuous line for monocular and stereo pipeline, black dashed for RGB-D. \cite{bescos2018dynaslam}}
\label{dyna}
\centering
\end{figure*}

\subsection{DynaSLAM}
DynaSLAM adds dynamic object detection and background in-painting to ORB-SLAM2. It is capable of detecting the moving objects either by multi-view geometry, deep learning or both, working in either monocular, stereo or RGB-D configuration \cite{bescos2018dynaslam}.
Fig. \ref{dyna} shows the DynaSLAM overview. The frames pass through a convolutional neural network that does pixel-wise segmentation of the dynamic content based on a pre-trained Mask R-CNN \cite{matterport_maskrcnn_2017} model containing people, vehicles, etc. In case of and RGB-D configuration that is combined by the proposed multi-view geometry motion segmentation that manages to detect the motions of inanimate objects like a book in a person's hand or a chair, further improving the segmentation of dynamic objects.
\begin{figure}[!th]
\includegraphics[width=\columnwidth]{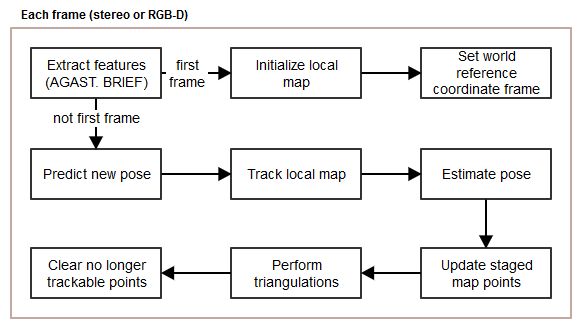}
\caption{A high-level overview of LVT SLAM algorithm.}
\label{lvtfigure}
\centering
\end{figure}
\subsection{Lightweight visual odometry}
Lightweight visual odometry (LVT) is a feature-based system that is compatible with both stereo and RGB-D sensors. Its use of a transient local map enables it to achieve similar estimation accuracies to common V-SLAM systems.  As it is designed to operate in real-time with low computational overhead and memory requirement it can be run on embedded single-board computes such as the Raspberry PI 3 and the ODROID XU4. To achieve that LVT uses corner-like features that are fast to compute, more precisely the adaptive and generic accelerated segment test (AGAST) \cite{mair2010adaptive} corner detector is used which is also an improvement of FAST \cite{rosten2006machine} corner detector. Additionally, BRIEF descriptors are uses. 
It also tackles the problem of distributing the detected features across the whole image, as a dense concentration of features in one region can lead to poor results. To do so the image is split into cells in which the features will be extracted from. These features are suppressed based on their corner strength using a technique known as adaptive non-maximal suppression \cite{brown2005multi} leaving only the local maxima in the neighborhood \cite{aladem2018lightweight}. A high-level overview of the system is shown in figure \ref{lvtfigure}.

\subsection{Stereo Direct Sparse Odometry}
Stereo Direct Sparse Odometry (SDSO) is a real-time visual odometry estimation of large-scale environments from stereo cameras. It is a visual odometry formulation that combines a fully direct probabilistic model with consistent, joint optimization of all model parameters, geometry and camera motion. Combining static stereo with multi-view stereo gives them several advantages over each of the separate ones. Unlike other direct methods, it jointly optimize for camera intrinsics, camera extrinsics, and the depth values, effectively performing the equivalent of windows sparse bundle adjustment \cite{wang2017stereo}.

\subsection{ProSLAM}
ProSLAM is a complete stereo visual SLAM system that combines well-known techniques.\cite{schlegel2018proslam} It uses a feature-based approach and with the use of the known geometry of the stereo cameras, it determines the 3D position of the tracked points. Tracked points that appear on multiple subsequent frames are grouped to form landmarks of which with a portion of trajectory are grouped into local maps and the local maps themselves are arranged in a pose graph. The pose graph allows the detection of know locations and adjusting the corresponding local map. 

\section{BENCHMARKING}
This section details the overall benchmarking methodology. It introduces the different datasets used, the setup where the tests were conducted, the results of the experiments, as well as the issues that arose. 

\subsection{Datasets}
The evaluations are performed on publicly available standard benchmark datasets with ground truth. We focus on choosing the different environments from them to compare the results depending on the setting.

\subsubsection{KITTI \cite{Geiger2013IJRR}}
The KITTI Vision Benchmark Suite is one of the go-to datasets to test on for any SLAM method. It provides 22 stereo sequences in a lossless png format. Half of which are provided with ground truth trajectories for training. The sequences are recorded from a driving car, motion patterns are limited to forward motion and environment to urban streets. Raw sensor measurements of calibration datasets are not available and the images are rectified. The ground truth contains GPS-INS poses for all frames. 

\subsubsection{EuRoC MAV \cite{Burri25012016}}
11 sequences from a MAV in three different indoor environments in the Robot Operating System (ROS) bag or Autonomous System Lab (ASL) dataset format. The dataset contains stereo images, synchronized IMU measurements, and accurate motion and structure ground truth per frame. It also provides the raw sensor data and a calibration dataset. 

\subsubsection{Considered dataset}
TUM Monocular \cite{engel2016monodataset} containing 50 real-world sequences recorded in different environments - from narrow indoor corridors to wide outdoor scenes. Dataset contains monocular photometrically calibrated images, exposure times for each frame and ground truth. Unfortunately, it was impossible to use that dataset because of two reasons. The main issue is that the ground truth is only partial, which made it impossible to get a reliable evaluation. Also, most of the methods tested rely on stereo vision.

\begin{table}[ht]
\caption{Datasets and sequences overview}
\begin{adjustbox}{width=\columnwidth,center}
\begin{tabular}{|l|l|c|c|c|}
\hline
Dataset - Seq      & Setting           & Frames & Resolution & Length (m)   \\ \hline
KITTI - 00         & Residential       & 4541   & 1241x376  & 3724.187     \\ \hline
KITTI - 01         & Highway           & 1101   & 1241x376  & 2453.203     \\ \hline
KITTI - 03         & Suburb            &  801   & 1241x376  & 560.888  \\ \hline
KITTI - 09         & Residential       & 1591   & 1241x376  & 1705.051     \\ \hline
EuRoC - MH\_01     & Indoor            & 3682   &  752x480  & 80.626     \\ \hline
EuRoC - V1\_01     & Indoor            & 2912   &  752x480  & 58.592     \\ \hline

\end{tabular}
\centering
\end{adjustbox}
\end{table}

\subsection{Setup}
The evaluations were ran on a portable computer with the following specs:
\begin{itemize}
    \item Intel(R) Core(TM) i7-8550U CPU (4 cores, 1.8GHz, Turbo 4.0GHz, 8MB cache), 16GB of RAM
\end{itemize}

To evaluate the odometry produced by the methods, a tool named \textit{evo} \cite{grupp2017evo} is used. It is a python package that supports handling, evaluating and comparing the trajectory outputs of odometry and SLAM algorithms. It supports the previously introduced datasets like KITTI and EuRoC MAV out of the box, as well as TUM Monocular \cite{engel2016monodataset} datasets and ROS bag files with certain message types. The built-in plotting provides excellent visuals and it also provides an implementation of SE(3) Umeyama alignment and scale that is usually required for monocular SLAMs, where calculating the scale is very difficult.

We will be evaluating the error in translation and rotation. The numbers that we will be looking at are the following; max, mean, median, min, rmse (root-mean-square error), sse (error sum of squares) and std (standard deviation). All sequences were run five times and the average of the runs was taken as the final result. 
For the better comparison of the results, the evaluation was done without modification and with alignment applied.

\subsection{Difficulties}
The V-SLAMS reviewed all have open-source code but it does not mean that getting them to work is an easy task. The level of documentation is very different and some of the cases require specific versions of third-party libraries. Also, the repositories are not well maintained. The compilation of the methods was not too complicated, the same can not be said for either; replicating the results on reference datasets or getting the algorithm to perform on a completely new dataset that was not covered. Due to the poor documentation running them on non-reference data proved to be complicated. Sample reasons being the dataset format, camera calibration, difference in input resolution, rectification, method parameters, missing binaries, etc. 

\subsection{Evaluation}
\begin{figure}[ht]
\includegraphics[width=\columnwidth]{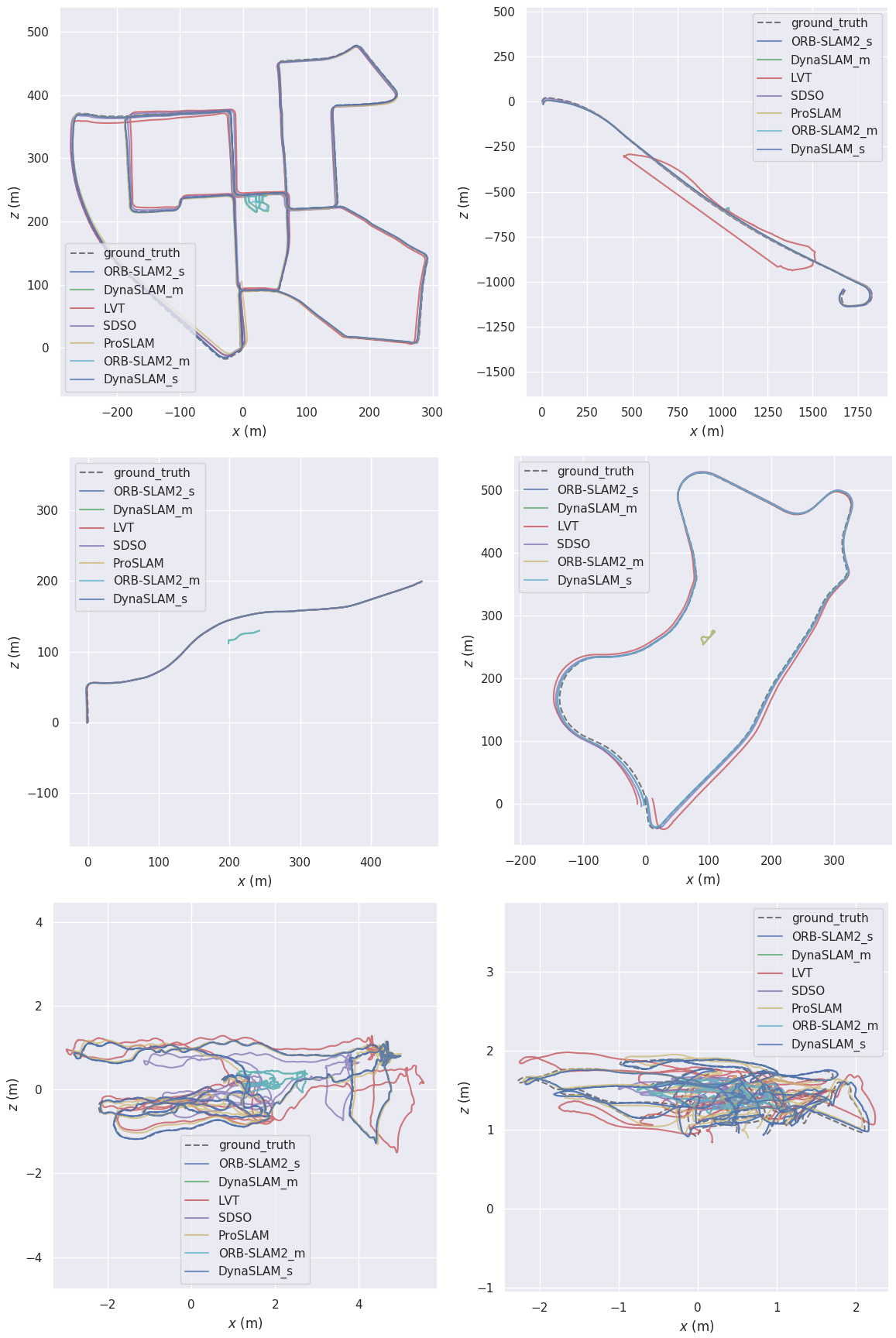}
\caption{Trajectories overlaid with the ground truth for each sequence, alignment applied. From top left: kitti-00, kitti-01, kitti-03, kitti-09, EuRoC-MH\_01, EuRoC-V1\_01}
\label{merged}
\centering
\end{figure}

Table \ref{results} reflects the evaluation results. Values "-" mean that we did not manage to run the method on that sequence. We have included two sets of results, one with no alignment or scaling and another one with SE(3) Umeyama alignment with respect to the ground truth. Figure \ref{merged} shows all the methods overlaid with the ground truth per sequence, best results are in bold.

A glance shows that the best performing SLAM overall would be ORB-SLAM2 in a stereo configuration. Besides, as expected DynaSLAM, has very similar results to ORB-SLAM2, of which it is built on top of, in both monocular and stereo setups.
In the case of indoor sequences, the monocular versions ORB-SLAM2 and DynaSLAM outperform their stereo counterparts as well as the other methods. On other sequences, monocular methods perform worse, which is probably because of the common issue of monocular methods, inability to correctly predict the scale.

\begin{figure}[ht]
\includegraphics[width=\columnwidth]{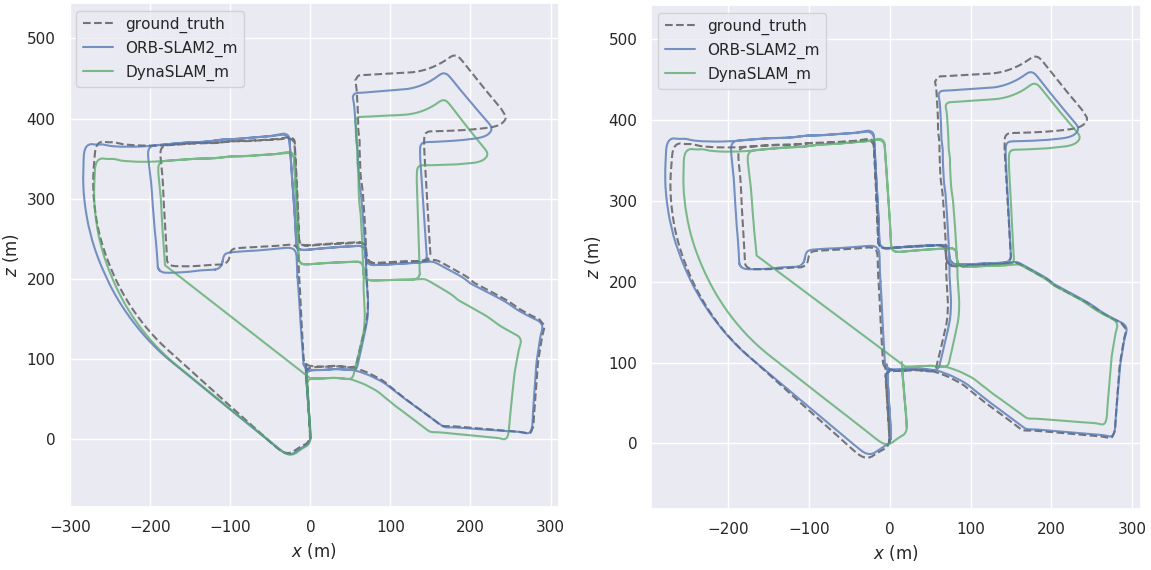}
\caption{Trajectory from ORB-SLAM2 and DynaSLAM monocular version outputs on kitti - 00 sequence with SE(3) scaling and alignment. Left only scaling, right scaling and alignment.}
\label{scaled}
\centering
\end{figure}

Figure \ref{scaled} displays the result of scaling and alignment on monocular methods outputs. It shows that the scaled-up version of the trajectory is very similar to the ground truth which clearly demonstrates the issue with scale prediction. However, applying alignment correction to stereo methods makes them significantly outperform monocular methods. 
From the results, it can be seen that the speed of the vehicle has a significant impact on the performance of SLAM systems. Methods across the board have higher translation error on the highway sequence (kitti-01). Method taking the largest hit compared to other methods was LVT.
It is also worth mentioning that the length of the sequence affects the overall accuracy of the methods. All methods have higher errors on longer sequences, which can be explained by the cumulative error building up.

\clearpage

\begin{table*}[ht!]
\caption{evaluation results}
\centering
\label{results}
\begin{adjustbox}{width=\textwidth}
\begin{tabular}{|c|l|l|l|l|l|l|l|l|}
\hline
\multicolumn{9}{|c|}{\begin{tabular}[c]{@{}c@{}}\textbf{No alignment or scaling - Translation (m) \textbf{/ Rotation (deg)}}\\\textbf{\textbf{Alignment applied - Translation (m) / Rotation (deg) }}\end{tabular}} \\ 
\hline
\textbf{Dataset} & \multicolumn{1}{c|}{\textbf{Method}} & \multicolumn{1}{c|}{\textbf{max}} & \multicolumn{1}{c|}{\textbf{mean}} & \multicolumn{1}{c|}{\textbf{median}} & \multicolumn{1}{c|}{\textbf{min}} & \multicolumn{1}{c|}{\textbf{rmse}} & \multicolumn{1}{c|}{\textbf{sse}} & \multicolumn{1}{c|}{\textbf{sdt}} \\ 
\hline
\multirow{7}{*}{\begin{tabular}[c]{@{}c@{}}\textbf{kitti - 00 }\\\textbf{Residental road }\end{tabular}} & LVT & \begin{tabular}[c]{@{}l@{}}39.087 / 0.233 \\16.812 / 0.162\end{tabular} & \begin{tabular}[c]{@{}l@{}}14.506 / 0.071 \\6.668 / 0.049\end{tabular} & \begin{tabular}[c]{@{}l@{}}11.918 / 0.071 \\6.224 / 0.051\end{tabular} & \begin{tabular}[c]{@{}l@{}}0 / 0 \\2.045 / 0.006\end{tabular} & \begin{tabular}[c]{@{}l@{}}17.063 / 0.077 \\7.443 / 0.054\end{tabular} & \begin{tabular}[c]{@{}l@{}}1322064.369 / 26.941 \\251572.494 / 13.238\end{tabular} & \begin{tabular}[c]{@{}l@{}}8.984 / 0.031 \\3.307 / 0.022\end{tabular} \\ 
\cline{2-9}
 & SDSO & \begin{tabular}[c]{@{}l@{}}22.64 / 0.234 \\7.446 / 0.17\end{tabular} & \begin{tabular}[c]{@{}l@{}}9.84 / 0.058 \\3.727 / 0.029\end{tabular} & \begin{tabular}[c]{@{}l@{}}9.079 / 0.057 \\3.744 / 0.025\end{tabular} & \begin{tabular}[c]{@{}l@{}}0 / 0 \\0.542 / 0.003\end{tabular} & \begin{tabular}[c]{@{}l@{}}11.063 / 0.063 \\4.09 / 0.034\end{tabular} & \begin{tabular}[c]{@{}l@{}}555764.873 / 17.798 \\75965.729 / 5.293\end{tabular} & \begin{tabular}[c]{@{}l@{}}5.056 / 0.023 \\1.684 / 0.018\end{tabular} \\ 
\cline{2-9}
 & ProSLAM & \begin{tabular}[c]{@{}l@{}}15.992 / 0.232 \\11.644 / 0.18\end{tabular} & \begin{tabular}[c]{@{}l@{}}8.34 / 0.056 \\2.629 / 0.027\end{tabular} & \begin{tabular}[c]{@{}l@{}}8.381 / 0.05 \\1.927 / 0.022\end{tabular} & \begin{tabular}[c]{@{}l@{}}0 / 0 \\0.796 / 0.006\end{tabular} & \begin{tabular}[c]{@{}l@{}}9.043 / 0.06\\3.472 / 0.03\end{tabular} & \begin{tabular}[c]{@{}l@{}}371360.961 / 16.601 \\54753.313 / 4.089\end{tabular} & \begin{tabular}[c]{@{}l@{}}3.496 / 0.022 \\2.269 / 0.014\end{tabular} \\ 
\cline{2-9}
 & ORB-mono & \begin{tabular}[c]{@{}l@{}}483.44 / 0.188 \\316.965 / 0.165\end{tabular} & \begin{tabular}[c]{@{}l@{}}260.234 / 0.036 \\162.888 / 0.018\end{tabular} & \begin{tabular}[c]{@{}l@{}}256.783 / 0.035 \\164.218 / 0.015\end{tabular} & \begin{tabular}[c]{@{}l@{}}0 / 0 \\11.63 / 0.001\end{tabular} & \begin{tabular}[c]{@{}l@{}}284.429 / 0.039 \\182.127 / 0.022\end{tabular} & \begin{tabular}[c]{@{}l@{}}367365240.039 / 6.994 \\150625601.066 / 2.22\end{tabular} & \begin{tabular}[c]{@{}l@{}}114.795 / 0.015 \\81.472 / 0.013\end{tabular} \\ 
\cline{2-9}
 & ORB-stereo & \begin{tabular}[c]{@{}l@{}}\textbf{12.263 / 0.194 }\\\textbf{3.551 / 0.166}\end{tabular} & \begin{tabular}[c]{@{}l@{}}\textbf{6.763 / 0.035 }\\\textbf{1.18 / 0.016}\end{tabular} & \begin{tabular}[c]{@{}l@{}}\textbf{6.623 / 0.035 }\\\textbf{1.081 / 0.014}\end{tabular} & \begin{tabular}[c]{@{}l@{}}\textbf{0 / 0 }\\\textbf{0.125 / 0.002}\end{tabular} & \begin{tabular}[c]{@{}l@{}}\textbf{7.48 / 0.037 }\\\textbf{1.318 / 0.019}\end{tabular} & \begin{tabular}[c]{@{}l@{}}\textbf{254091.314 / 6.333 }\\\textbf{7885.096 / 1.634}\end{tabular} & \begin{tabular}[c]{@{}l@{}}\textbf{3.197 / 0.012 }\\\textbf{0.586 / 0.011}\end{tabular} \\ 
\cline{2-9}
 & Dyna-mono & \begin{tabular}[c]{@{}l@{}}488.627 / 0.175 \\314.46 / 2.795\end{tabular} & \begin{tabular}[c]{@{}l@{}}263.341 / 0.037 \\161.26 / 0.207\end{tabular} & \begin{tabular}[c]{@{}l@{}}260.164 / 0.036 \\162.8 / 0.048\end{tabular} & \begin{tabular}[c]{@{}l@{}}0 / 0 \\11.824 / 0.003\end{tabular} & \begin{tabular}[c]{@{}l@{}}287.798 / 0.04 \\180.657 / 0.524\end{tabular} & \begin{tabular}[c]{@{}l@{}}376120608.959 / 7.104 \\148204214.918 / 1249.124\end{tabular} & \begin{tabular}[c]{@{}l@{}}116.101 / 0.013 \\81.438 / 0.482\end{tabular} \\ 
\cline{2-9}
 & Dyna-stereo & \begin{tabular}[c]{@{}l@{}}13.043 / 0.19\\3.593 / 0.163\end{tabular} & \begin{tabular}[c]{@{}l@{}}6.797 / 0.036 \\1.253 / 0.016\end{tabular} & \begin{tabular}[c]{@{}l@{}}6.542 / 0.034 \\1.173 / 0.014\end{tabular} & \begin{tabular}[c]{@{}l@{}}0 / 0 \\0.066 / 0.001\end{tabular} & \begin{tabular}[c]{@{}l@{}}7.518 / 0.038 \\1.391 / 0.019\end{tabular} & \begin{tabular}[c]{@{}l@{}}256624.656 / 6.391 \\8784.549 / 1.629\end{tabular} & \begin{tabular}[c]{@{}l@{}}3.212 / 0.012 \\0.603 / 0.011\end{tabular} \\ 
\hline \hline
\multirow{7}{*}{\begin{tabular}[c]{@{}c@{}}\textbf{kitti - 02 }\\\textbf{Highway road }\end{tabular}} & LVT & \begin{tabular}[c]{@{}l@{}}1994.865 / 1.42 \\1451.113 / 1.559\end{tabular} & \begin{tabular}[c]{@{}l@{}}1042.16 / 1.32 \\372.504 / 0.145\end{tabular} & \begin{tabular}[c]{@{}l@{}}1083.569 / 1.354 \\369.602 / 0.078\end{tabular} & \begin{tabular}[c]{@{}l@{}}0 / 0 \\16.192 / 0.015\end{tabular} & \begin{tabular}[c]{@{}l@{}}1213.969 / 1.334 \\459.755 / 0.291\end{tabular} & \begin{tabular}[c]{@{}l@{}}1622567829.638 / 1958.664 \\232723158.486 / 93.365\end{tabular} & \begin{tabular}[c]{@{}l@{}}622.595 / 0.194 \\269.472 / 0.252\end{tabular} \\ 
\cline{2-9}
 & SDSO & \begin{tabular}[c]{@{}l@{}}40.37 / 0.039 \\10.836 / 0.029\end{tabular} & \begin{tabular}[c]{@{}l@{}}23.929 / 0.022 \\4.072 / 0.013\end{tabular} & \begin{tabular}[c]{@{}l@{}}26.185 / 0.022 \\3.052 / 0.013\end{tabular} & \begin{tabular}[c]{@{}l@{}}0.001 / 0 \\\textbf{0.853 / 0.005}\end{tabular} & \begin{tabular}[c]{@{}l@{}}26.646 / 0.022 \\5.08 / 0.014\end{tabular} & \begin{tabular}[c]{@{}l@{}}781701.021 / 0.547 \\\textbf{28414.621 / 0.214}\end{tabular} & \begin{tabular}[c]{@{}l@{}}11.721 / 0.004 \\3.038 / 0.004\end{tabular} \\ 
\cline{2-9}
 & ProSLAM & \begin{tabular}[c]{@{}l@{}}48.876 / 0.359 \\46.256 / 0.359\end{tabular} & \begin{tabular}[c]{@{}l@{}}\textbf{12.558 / 0.043 }\\\textbf{9.287 / 0.04}\end{tabular} & \begin{tabular}[c]{@{}l@{}}\textbf{10.49 / 0.037 }\\\textbf{7.634 / 0.033}\end{tabular} & \begin{tabular}[c]{@{}l@{}}\textbf{0 / 0 }\\4.646 / 0.011\end{tabular} & \begin{tabular}[c]{@{}l@{}}\textbf{13.872 / 0.06 }\\10.507 / 0.057\end{tabular} & \begin{tabular}[c]{@{}l@{}}\textbf{211862.01 / 3.964} \\121544.87 / 3.595\end{tabular} & \begin{tabular}[c]{@{}l@{}}\textbf{5.893 / 0.042} \\4.913 / 0.041\end{tabular} \\ 
\cline{2-9}
 & ORB-mono & \begin{tabular}[c]{@{}l@{}}2119.869 / 2.828 \\1194.775 / 2.828\end{tabular} & \begin{tabular}[c]{@{}l@{}}1192.283 / 1.027 \\641.606 / 0.875\end{tabular} & \begin{tabular}[c]{@{}l@{}}1251.719 / 0.806 \\715.366 / 0.632\end{tabular} & \begin{tabular}[c]{@{}l@{}}0 / 0 \\23.561 / 0.321\end{tabular} & \begin{tabular}[c]{@{}l@{}}1391.002 / 1.235 \\721.542 / 1.112\end{tabular} & \begin{tabular}[c]{@{}l@{}}2130309352.695 / 1679.377 \\573205406.929 / 1360.23\end{tabular} & \begin{tabular}[c]{@{}l@{}}716.482 / 0.686 \\330.097 / 0.686\end{tabular} \\ 
\cline{2-9}
 & ORB-stereo & \begin{tabular}[c]{@{}l@{}}\textbf{30.347 / 0.066 }\\\textbf{16.78 / 0.049}\end{tabular} & \begin{tabular}[c]{@{}l@{}}17.229 / 0.036 \\9.449 / 0.029\end{tabular} & \begin{tabular}[c]{@{}l@{}}17.356 / 0.038 \\8.277 / 0.028\end{tabular} & \begin{tabular}[c]{@{}l@{}}0 / 0 \\6.359 / 0.012\end{tabular} & \begin{tabular}[c]{@{}l@{}}19.237 / 0.039 \\\textbf{9.883 / 0.03}\end{tabular} & \begin{tabular}[c]{@{}l@{}}407458.106 / 1.686 \\107542.076 / 0.981\end{tabular} & \begin{tabular}[c]{@{}l@{}}8.558 / 0.016 \\\textbf{2.896 / 0.009}\end{tabular} \\ 
\cline{2-9}
 & Dyna-mono & \begin{tabular}[c]{@{}l@{}}2141.068 / 2.857 \\1206.722 / 2.857\end{tabular} & \begin{tabular}[c]{@{}l@{}}1204.206 / 1.037 \\648.022 / 0.883\end{tabular} & \begin{tabular}[c]{@{}l@{}}1264.236 / 0.814 \\722.52 / 0.638\end{tabular} & \begin{tabular}[c]{@{}l@{}}0 / 0 \\23.797 / 0.324\end{tabular} & \begin{tabular}[c]{@{}l@{}}1404.912 / 1.247 \\728.757 / 1.123\end{tabular} & \begin{tabular}[c]{@{}l@{}}2151612446.222 / 1696.171 \\578937460.999 / 1373.833\end{tabular} & \begin{tabular}[c]{@{}l@{}}723.647 / 0.693 \\333.398 / 0.693\end{tabular} \\ 
\cline{2-9}
 & Dyna-stereo & \begin{tabular}[c]{@{}l@{}}30.65 / 0.065 \\16.947 / 0.05\end{tabular} & \begin{tabular}[c]{@{}l@{}}17.401 / 0.035 \\9.544 / 0.029\end{tabular} & \begin{tabular}[c]{@{}l@{}}17.529 / 0.038 \\8.359 / 0.028\end{tabular} & \begin{tabular}[c]{@{}l@{}}0 / 0\\6.423 / 0.012\end{tabular} & \begin{tabular}[c]{@{}l@{}}19.43 / 0.038 \\9.982 / 0.03\end{tabular} & \begin{tabular}[c]{@{}l@{}}411532.687 / 1.656 \\108617.497 / 0.991\end{tabular} & \begin{tabular}[c]{@{}l@{}}8.644 / 0.016 \\2.925 / 0.009\end{tabular} \\ 
\hline \hline
\multirow{7}{*}{\begin{tabular}[c]{@{}c@{}}\textbf{kitti - 03 }\\\textbf{Suburb road }\end{tabular}} & LVT & \begin{tabular}[c]{@{}l@{}}5.69 / 0.029\\\textbf{0.835 / 0.047}\end{tabular} & \begin{tabular}[c]{@{}l@{}}2.625 / 0.016 \\\textbf{0.466 / 0.037}\end{tabular} & \begin{tabular}[c]{@{}l@{}}2.49 / 0.017 \\\textbf{0.445 / 0.037}\end{tabular} & \begin{tabular}[c]{@{}l@{}}0 / 0 \\0.14 / 0.023\end{tabular} & \begin{tabular}[c]{@{}l@{}}3.128 / 0.017 \\\textbf{0.494 / 0.037}\end{tabular} & \begin{tabular}[c]{@{}l@{}}7834.945 / 0.238 \\\textbf{195.758 / 1.103}\end{tabular} & \begin{tabular}[c]{@{}l@{}}1.7 / 0.007 \\\textbf{0.165 / 0.004}\end{tabular} \\ 
\cline{2-9}
 & SDSO & \begin{tabular}[c]{@{}l@{}}5.163 / 0.017 \\2.387 / 0.022\end{tabular} & \begin{tabular}[c]{@{}l@{}}2.453 / 0.012 \\1.216 / 0.013\end{tabular} & \begin{tabular}[c]{@{}l@{}}1.915 / 0.013 \\0.965 / 0.014\end{tabular} & \begin{tabular}[c]{@{}l@{}}0 / 0 \\0.023 / 0.003\end{tabular} & \begin{tabular}[c]{@{}l@{}}2.906 / 0.012 \\1.38 / 0.014\end{tabular} & \begin{tabular}[c]{@{}l@{}}6765.145 / 0.124 \\1525.731 / 0.148\end{tabular} & \begin{tabular}[c]{@{}l@{}}1.558 / 0.004 \\0.652 / 0.005\end{tabular} \\ 
\cline{2-9}
 & ProSLAM & \begin{tabular}[c]{@{}l@{}}5.304 / 0.022 \\2.492 / 0.022\end{tabular} & \begin{tabular}[c]{@{}l@{}}2.479 / 0.013 \\1.184 / 0.013\end{tabular} & \begin{tabular}[c]{@{}l@{}}2.031 / 0.014 \\1.037 / 0.014\end{tabular} & \begin{tabular}[c]{@{}l@{}}0 / 0 \\0.144 / 0.003\end{tabular} & \begin{tabular}[c]{@{}l@{}}2.946 / 0.014 \\1.347 / 0.014\end{tabular} & \begin{tabular}[c]{@{}l@{}}6954.067 / 0.166 \\1453.871 / 0.148\end{tabular} & \begin{tabular}[c]{@{}l@{}}1.593 / 0.006 \\0.644 / 0.005\end{tabular} \\ 
\cline{2-9}
 & ORB-mono & \begin{tabular}[c]{@{}l@{}}464.732 / 0.014 \\239.052 / 0.017\end{tabular} & \begin{tabular}[c]{@{}l@{}}232.635 / 0.008 \\137.462 / 0.009\end{tabular} & \begin{tabular}[c]{@{}l@{}}227.564 / 0.009 \\140.929 / 0.009\end{tabular} & \begin{tabular}[c]{@{}l@{}}0.082 / 0 \\24.046 / 0.001\end{tabular} & \begin{tabular}[c]{@{}l@{}}275.497 / 0.008 \\154.01 / 0.01\end{tabular} & \begin{tabular}[c]{@{}l@{}}60794773.07 / 0.056 \\18998993.677 / 0.074\end{tabular} & \begin{tabular}[c]{@{}l@{}}147.579 / 0.003 \\69.45 / 0.003\end{tabular} \\ 
\cline{2-9}
 & ORB-stereo & \begin{tabular}[c]{@{}l@{}}\textbf{2.831 / 0.017 }\\1.396 / 0.017\end{tabular} & \begin{tabular}[c]{@{}l@{}}\textbf{1.492 / 0.008 }\\0.723 / 0.009\end{tabular} & \begin{tabular}[c]{@{}l@{}}\textbf{1.274 / 0.008 }\\0.61 / 0.009\end{tabular} & \begin{tabular}[c]{@{}l@{}}\textbf{0 / 0 }\\\textbf{0.006 / 0.002}\end{tabular} & \begin{tabular}[c]{@{}l@{}}\textbf{1.672 / 0.008 }\\0.822 / 0.01\end{tabular} & \begin{tabular}[c]{@{}l@{}}\textbf{2240.434 / 0.057 }\\541.423 / 0.079\end{tabular} & \begin{tabular}[c]{@{}l@{}}\textbf{0.755 / 0.003 }\\0.392 / 0.003\end{tabular} \\ 
\cline{2-9}
 & Dyna-mono & \begin{tabular}[c]{@{}l@{}}520.5 / 0.015 \\267.738 / 0.019\end{tabular} & \begin{tabular}[c]{@{}l@{}}260.551 / 0.009 \\153.957 / 0.01\end{tabular} & \begin{tabular}[c]{@{}l@{}}254.871 / 0.009 \\157.841 / 0.01\end{tabular} & \begin{tabular}[c]{@{}l@{}}0.092 / 0 \\26.932 / 0.002\end{tabular} & \begin{tabular}[c]{@{}l@{}}308.557 / 0.009 \\172.491 / 0.011\end{tabular} & \begin{tabular}[c]{@{}l@{}}68090145.838 / 0.062 \\21278872.918 / 0.083\end{tabular} & \begin{tabular}[c]{@{}l@{}}165.289 / 0.003 \\77.784 / 0.003\end{tabular} \\ 
\cline{2-9}
 & Dyna-stereo & \begin{tabular}[c]{@{}l@{}}3.057 / 0.019 \\1.508 / 0.019\end{tabular} & \begin{tabular}[c]{@{}l@{}}1.612 / 0.009 \\0.78 / 0.01\end{tabular} & \begin{tabular}[c]{@{}l@{}}1.376 / 0.008 \\0.659 / 0.01\end{tabular} & \begin{tabular}[c]{@{}l@{}}0 / 0 \\0.007 / 0.002\end{tabular} & \begin{tabular}[c]{@{}l@{}}1.806 / 0.009 \\0.888 / 0.011\end{tabular} & \begin{tabular}[c]{@{}l@{}}2419.669 / 0.063 \\584.737 / 0.086\end{tabular} & \begin{tabular}[c]{@{}l@{}}0.815 / 0.004 \\0.424 / 0.003\end{tabular} \\ 
\hline \hline
\multirow{7}{*}{\begin{tabular}[c]{@{}c@{}}\textbf{kitti - 09}\\\textbf{Residental road }\end{tabular}} & LVT & \begin{tabular}[c]{@{}l@{}}35.731 / 0.093 \\18.309 / 0.126\end{tabular} & \begin{tabular}[c]{@{}l@{}}12.18 / 0.045 \\7.347 / 0.07\end{tabular} & \begin{tabular}[c]{@{}l@{}}6.18 / 0.043 \\5.565 / 0.071\end{tabular} & \begin{tabular}[c]{@{}l@{}}0 / 0 \\0.427 / 0.02\end{tabular} & \begin{tabular}[c]{@{}l@{}}16.41 / 0.051 \\8.892 / 0.075\end{tabular} & \begin{tabular}[c]{@{}l@{}}428449.319 / 4.187 \\125805.619 / 9.064\end{tabular} & \begin{tabular}[c]{@{}l@{}}10.997 / 0.024 \\5.009 / 0.027\end{tabular} \\ 
\cline{2-9}
 & SDSO & \begin{tabular}[c]{@{}l@{}}15.757 / 0.03 \\9.819 / 0.048\end{tabular} & \begin{tabular}[c]{@{}l@{}}8.287 / 0.016 \\3.63 / 0.026\end{tabular} & \begin{tabular}[c]{@{}l@{}}8.639 / 0.017 \\2.903 / 0.024\end{tabular} & \begin{tabular}[c]{@{}l@{}}0 / 0 \\0.796 / 0.011\end{tabular} & \begin{tabular}[c]{@{}l@{}}9.389 / 0.017 \\4.311 / 0.027\end{tabular} & \begin{tabular}[c]{@{}l@{}}140241.159 / 0.461 \\29573.237 / 1.17\end{tabular} & \begin{tabular}[c]{@{}l@{}}4.413 / 0.007 \\2.326 / 0.009\end{tabular} \\ 
\cline{2-9}
 & ProSLAM & \begin{tabular}[c]{@{}l@{}}- / -~ \\- / -\end{tabular} & \begin{tabular}[c]{@{}l@{}}- / -~ \\- / -\end{tabular} & \begin{tabular}[c]{@{}l@{}}- / -~ \\- / -\end{tabular} & \begin{tabular}[c]{@{}l@{}}- / -~ \\- / -\end{tabular} & \begin{tabular}[c]{@{}l@{}}0 / 0 \\0 / 0\end{tabular} & \begin{tabular}[c]{@{}l@{}}- / -~ \\- / -\end{tabular} & \begin{tabular}[c]{@{}l@{}}- / -~ \\- / -\end{tabular} \\ 
\cline{2-9}
 & ORB-mono & \begin{tabular}[c]{@{}l@{}}561.532 / 2.828 \\329.054 / 2.828\end{tabular} & \begin{tabular}[c]{@{}l@{}}301.753 / 0.856 \\209.343 / 1.095\end{tabular} & \begin{tabular}[c]{@{}l@{}}293.249 / 0.033 \\227.949 / 0.711\end{tabular} & \begin{tabular}[c]{@{}l@{}}0.045 / 0.001 \\57.4 / 0.146\end{tabular} & \begin{tabular}[c]{@{}l@{}}348.254 / 1.315 \\220.132 / 1.238\end{tabular} & \begin{tabular}[c]{@{}l@{}}192957464.167 / 2750.501 \\77096805.468 / 2438.31\end{tabular} & \begin{tabular}[c]{@{}l@{}}173.856 / 0.998 \\68.071 / 0.577\end{tabular} \\ 
\cline{2-9}
 & ORB-stereo & \begin{tabular}[c]{@{}l@{}}\textbf{12.736 / 0.048 }\\\textbf{6.708 / 0.059}\end{tabular} & \begin{tabular}[c]{@{}l@{}}\textbf{5.256 / 0.023 }\\\textbf{2.522 / 0.027}\end{tabular} & \begin{tabular}[c]{@{}l@{}}\textbf{4.305 / 0.023 }\\\textbf{1.971 / 0.027}\end{tabular} & \begin{tabular}[c]{@{}l@{}}\textbf{0 / 0 }\\\textbf{0.295 / 0.002}\end{tabular} & \begin{tabular}[c]{@{}l@{}}\textbf{6.331 / 0.026 }\\\textbf{3.002 / 0.03}\end{tabular} & \begin{tabular}[c]{@{}l@{}}\textbf{63767.702 / 1.05 }\\\textbf{14333.689 / 1.393}\end{tabular} & \begin{tabular}[c]{@{}l@{}}\textbf{3.529 / 0.011 }\\\textbf{1.627 / 0.012}\end{tabular} \\ 
\cline{2-9}
 & Dyna-mono & \begin{tabular}[c]{@{}l@{}}634.531 / 3.026 \\371.831 / 3.196\end{tabular} & \begin{tabular}[c]{@{}l@{}}340.981 / 0.916 \\236.557 / 1.237\end{tabular} & \begin{tabular}[c]{@{}l@{}}331.371 / 0.036 \\257.582 / 0.804\end{tabular} & \begin{tabular}[c]{@{}l@{}}0.051 / 0.001 \\64.862 / 0.165\end{tabular} & \begin{tabular}[c]{@{}l@{}}393.527 / 1.407 \\248.749 / 1.399\end{tabular} & \begin{tabular}[c]{@{}l@{}}218041934.508 / 2943.036 \\87119390.179 / 2755.291\end{tabular} & \begin{tabular}[c]{@{}l@{}}196.457 / 1.068 \\76.92 / 0.652\end{tabular} \\ 
\cline{2-9}
 & Dyna-stereo & \begin{tabular}[c]{@{}l@{}}14.009 / 0.052 \\7.379 / 0.065\end{tabular} & \begin{tabular}[c]{@{}l@{}}5.782 / 0.025 \\2.774 / 0.03\end{tabular} & \begin{tabular}[c]{@{}l@{}}4.736 / 0.025 \\2.168 / 0.03\end{tabular} & \begin{tabular}[c]{@{}l@{}}0 / 0 \\0.324 / 0.002\end{tabular} & \begin{tabular}[c]{@{}l@{}}6.964 / 0.028 \\3.302 / 0.033\end{tabular} & \begin{tabular}[c]{@{}l@{}}70144.472 / 1.144 \\15767.058 / 1.532\end{tabular} & \begin{tabular}[c]{@{}l@{}}3.881 / 0.012 \\1.79 / 0.013\end{tabular} \\ 
\hline \hline
\multirow{7}{*}{\begin{tabular}[c]{@{}c@{}}\textbf{EuRoC }\\\textbf{MH\_01 }\\\textbf{Indoor }\end{tabular}} & LVT & \begin{tabular}[c]{@{}l@{}}10.77 / 2.828 \\1.037 / 2.012\end{tabular} & \begin{tabular}[c]{@{}l@{}}6.082 / 2.623 \\0.392 / 1.932\end{tabular} & \begin{tabular}[c]{@{}l@{}}5.358 / 2.652 \\0.367 / 1.934\end{tabular} & \begin{tabular}[c]{@{}l@{}}2.722 / 2.007 \\0.059 / 1.892\end{tabular} & \begin{tabular}[c]{@{}l@{}}6.317 / 2.631 \\0.428 / 1.932\end{tabular} & \begin{tabular}[c]{@{}l@{}}145193.144 / 25181.646 \\667.294 / 13581.695\end{tabular} & \begin{tabular}[c]{@{}l@{}}\textbf{1.708 / 0.208 }\\0.172 / 0.027\end{tabular} \\ 
\cline{2-9}
 & SDSO & \begin{tabular}[c]{@{}l@{}}14.128 / 2.828 \\2.614 / 1.995\end{tabular} & \begin{tabular}[c]{@{}l@{}}6.918 / 2.438 \\1.469 / 1.97\end{tabular} & \begin{tabular}[c]{@{}l@{}}5.313 / 2.476 \\1.693 / 1.969\end{tabular} & \begin{tabular}[c]{@{}l@{}}2.36 / 1.668 \\0.327 / 1.958\end{tabular} & \begin{tabular}[c]{@{}l@{}}7.591 / 2.456 \\1.57 / 1.97\end{tabular} & \begin{tabular}[c]{@{}l@{}}\textbf{40396.22 / 4228.136 }\\1728.912 / 2719.279\end{tabular} & \begin{tabular}[c]{@{}l@{}}3.125 / 0.293 \\0.554 / 0.008\end{tabular} \\ 
\cline{2-9}
 & ProSLAM & \begin{tabular}[c]{@{}l@{}}17.577 / 2.828 \\0.204 / 2.043\end{tabular} & \begin{tabular}[c]{@{}l@{}}7.763 / 2.415 \\0.096 / 1.976\end{tabular} & \begin{tabular}[c]{@{}l@{}}5.642 / 2.383 \\0.1 / 1.979\end{tabular} & \begin{tabular}[c]{@{}l@{}}\textbf{2.188 / 1.67 }\\0.02 / 1.93\end{tabular} & \begin{tabular}[c]{@{}l@{}}8.906 / 2.432 \\0.103 / 1.976\end{tabular} & \begin{tabular}[c]{@{}l@{}}288550.648 / 21522.311 \\38.893 / 14209.087\end{tabular} & \begin{tabular}[c]{@{}l@{}}4.365 / 0.288 \\0.038 / 0.02\end{tabular} \\ 
\cline{2-9}
 & ORB-mono & \begin{tabular}[c]{@{}l@{}}\textbf{10.553 / 2.828 }\\5.884 / 2.011\end{tabular} & \begin{tabular}[c]{@{}l@{}}\textbf{5.596 / 2.391 }\\3.113 / 1.986\end{tabular} & \begin{tabular}[c]{@{}l@{}}\textbf{4.934 / 2.377 }\\3.469 / 1.984\end{tabular} & \begin{tabular}[c]{@{}l@{}}2.598 / 1.659 \\0.738 / 1.969\end{tabular} & \begin{tabular}[c]{@{}l@{}}\textbf{5.889 / 2.41} \\3.381 / 1.986\end{tabular} & \begin{tabular}[c]{@{}l@{}}126164.17 / 21134.679 \\41583.365 / 14342.351\end{tabular} & \begin{tabular}[c]{@{}l@{}}1.834 / 0.303 \\1.319 / 0.008\end{tabular} \\ 
\cline{2-9}
 & ORB-stereo & \begin{tabular}[c]{@{}l@{}}17.753 / 2.828 \\\textbf{0.091 / 2.008}\end{tabular} & \begin{tabular}[c]{@{}l@{}}7.832 / 2.394 \\\textbf{0.031 / 1.984}\end{tabular} & \begin{tabular}[c]{@{}l@{}}5.702 / 2.378 \\\textbf{0.024 / 1.983}\end{tabular} & \begin{tabular}[c]{@{}l@{}}2.237 / 1.664 \\\textbf{0.002 / 1.97}\end{tabular} & \begin{tabular}[c]{@{}l@{}}9 / 2.413 \\\textbf{0.037 / 1.984}\end{tabular} & \begin{tabular}[c]{@{}l@{}}294659.208 / 21180.483 \\\textbf{4.908 / 14320.687}\end{tabular} & \begin{tabular}[c]{@{}l@{}}4.433 / 0.302 \\\textbf{0.02 / 0.008}\end{tabular} \\ 
\cline{2-9}
 & Dyna-mono & \begin{tabular}[c]{@{}l@{}}10.975 / 2.828\\6.119 / 2.052\end{tabular} & \begin{tabular}[c]{@{}l@{}}5.82 / 2.396\\3.238 / 2.025\end{tabular} & \begin{tabular}[c]{@{}l@{}}5.132 / 2.376\\3.607 / 2.024\end{tabular} & \begin{tabular}[c]{@{}l@{}}2.702 / 1.652\\0.768 / 2.008\end{tabular} & \begin{tabular}[c]{@{}l@{}}6.124 / 2.413\\3.516 / 2.025\end{tabular} & \begin{tabular}[c]{@{}l@{}}131210.737 / 21232.674\\43246.699 / 14629.198\end{tabular} & \begin{tabular}[c]{@{}l@{}}1.907 / 0.305\\1.371 / 0.009\end{tabular} \\ 
\cline{2-9}
 & Dyna-stereo & \begin{tabular}[c]{@{}l@{}}19.173 / 2.828\\0.098 / 2.108\end{tabular} & \begin{tabular}[c]{@{}l@{}}8.459 / 2.395\\0.033 / 2.083\end{tabular} & \begin{tabular}[c]{@{}l@{}}6.158 / 2.388\\0.026 / 2.082\end{tabular} & \begin{tabular}[c]{@{}l@{}}2.416 / 1.673\\0.002 / 2.068\end{tabular} & \begin{tabular}[c]{@{}l@{}}9.72 / 2.424\\0.04 / 2.083\end{tabular} & \begin{tabular}[c]{@{}l@{}}318231.944 / 21280.483\\5.3 / 15036.721\end{tabular} & \begin{tabular}[c]{@{}l@{}}4.788 / 0.303\\0.021 / 0.008\end{tabular} \\ 
\hline \hline
\multirow{7}{*}{\begin{tabular}[c]{@{}c@{}}\textbf{EuRoC }\\\textbf{V1\_01 }\\\textbf{Indoor-Feature rich }\end{tabular}} & LVT & \begin{tabular}[c]{@{}l@{}}7.962 / 2.828 \\0.671 / 2.013\end{tabular} & \begin{tabular}[c]{@{}l@{}}4.018 / 2.67 \\0.328 / 1.933\end{tabular} & \begin{tabular}[c]{@{}l@{}}3.737 / 2.696 \\0.31 / 1.947\end{tabular} & \begin{tabular}[c]{@{}l@{}}1.497 / 2.087 \\0.022 / 1.856\end{tabular} & \begin{tabular}[c]{@{}l@{}}4.314 / 2.674 \\0.362 / 1.934\end{tabular} & \begin{tabular}[c]{@{}l@{}}53444.412 / 20542.518 \\377.164 / 10738.585\end{tabular} & \begin{tabular}[c]{@{}l@{}}1.571 / 0.161 \\0.154 / 0.044\end{tabular} \\ 
\cline{2-9}
 & SDSO & \begin{tabular}[c]{@{}l@{}}5.797 / 2.828 \\1.7 / 1.939\end{tabular} & \begin{tabular}[c]{@{}l@{}}2.943 / 2.695 \\0.893 / 1.913\end{tabular} & \begin{tabular}[c]{@{}l@{}}2.849 / 2.807 \\0.908 / 1.911\end{tabular} & \begin{tabular}[c]{@{}l@{}}\textbf{1.229 / 1.698 }\\0.142 / 1.896\end{tabular} & \begin{tabular}[c]{@{}l@{}}3.075 / 2.705 \\0.971 / 1.913\end{tabular} & \begin{tabular}[c]{@{}l@{}}\textbf{6664.349 / 5159.248} \\665.018 / 2580.749\end{tabular} & \begin{tabular}[c]{@{}l@{}}0.888 / 0.229 \\0.382 / 0.008\end{tabular} \\ 
\cline{2-9}
 & ProSLAM & \begin{tabular}[c]{@{}l@{}}8.29 / 2.828 \\0.24 / 1.981\end{tabular} & \begin{tabular}[c]{@{}l@{}}3.735 / 2.688 \\0.118 / 1.909\end{tabular} & \begin{tabular}[c]{@{}l@{}}3.594 / 2.808 \\0.105 / 1.906\end{tabular} & \begin{tabular}[c]{@{}l@{}}1.372 / 1.676\\0.038 / 1.878\end{tabular} & \begin{tabular}[c]{@{}l@{}}3.964 / 2.699 \\0.129 / 1.909\end{tabular} & \begin{tabular}[c]{@{}l@{}}45129.717 / 20927.634 \\47.651 / 10468.97\end{tabular} & \begin{tabular}[c]{@{}l@{}}1.327 / 0.246 \\0.051 / 0.015\end{tabular} \\ 
\cline{2-9}
 & ORB-mono & \begin{tabular}[c]{@{}l@{}}\textbf{5.509 / 2.828} \\1.94 / 1.93\end{tabular} & \begin{tabular}[c]{@{}l@{}}\textbf{2.785 / 2.675 }\\0.939 / 1.913\end{tabular} & \begin{tabular}[c]{@{}l@{}}\textbf{2.685 / 2.814 }\\0.935 / 1.913\end{tabular} & \begin{tabular}[c]{@{}l@{}}1.244 / 1.641 \\0.114 / 1.897\end{tabular} & \begin{tabular}[c]{@{}l@{}}\textbf{2.909 / 2.688 }\\1.031 / 1.913\end{tabular} & \begin{tabular}[c]{@{}l@{}}23458.273 / 20035.351 \\2944.79 / 10150.029\end{tabular} & \begin{tabular}[c]{@{}l@{}}\textbf{0.839 / 0.262 }\\0.425 / 0.007\end{tabular} \\ 
\cline{2-9}
 & ORB-stereo & \begin{tabular}[c]{@{}l@{}}8.282 / 2.828 \\\textbf{0.169 / 1.931}\end{tabular} & \begin{tabular}[c]{@{}l@{}}3.782 / 2.688 \\\textbf{0.082 / 1.913}\end{tabular} & \begin{tabular}[c]{@{}l@{}}3.649 / 2.807 \\\textbf{0.075 / 1.913}\end{tabular} & \begin{tabular}[c]{@{}l@{}}1.381 / 1.702 \\\textbf{0.016 / 1.895}\end{tabular} & \begin{tabular}[c]{@{}l@{}}4.002 / 2.699 \\\textbf{0.087 / 1.913}\end{tabular} & \begin{tabular}[c]{@{}l@{}}46004.765 / 20924.218 \\\textbf{21.97 / 10513.553}\end{tabular} & \begin{tabular}[c]{@{}l@{}}1.308 / 0.244 \\\textbf{0.031 / 0.008}\end{tabular} \\ 
\cline{2-9}
 & Dyna-mono & \begin{tabular}[c]{@{}l@{}}5.674 / 2.828 \\1.998 / 1.988\end{tabular} & \begin{tabular}[c]{@{}l@{}}2.868 / 2.702 \\0.967 / 1.971\end{tabular} & \begin{tabular}[c]{@{}l@{}}2.765 / 2.842 \\0.963 / 1.97\end{tabular} & \begin{tabular}[c]{@{}l@{}}1.282 / 1.658 \\0.118 / 1.954\end{tabular} & \begin{tabular}[c]{@{}l@{}}2.996 / 2.715 \\1.061 / 1.971\end{tabular} & \begin{tabular}[c]{@{}l@{}}24162.021 / 20235.704 \\3033.133 / 10454.529\end{tabular} & \begin{tabular}[c]{@{}l@{}}0.864 / 0.265 \\0.438 / 0.007\end{tabular} \\ 
\cline{2-9}
 & Dyna-stereo & \begin{tabular}[c]{@{}l@{}}8.696 / 2.828 \\0.177 / 1.97\end{tabular} & \begin{tabular}[c]{@{}l@{}}3.971 / 2.715 \\0.086 / 1.952\end{tabular} & \begin{tabular}[c]{@{}l@{}}3.832 / 2.835 \\0.079 / 1.951\end{tabular} & \begin{tabular}[c]{@{}l@{}}1.45 / 1.719 \\0.016 / 1.933\end{tabular} & \begin{tabular}[c]{@{}l@{}}4.202 / 2.726 \\0.092 / 1.952\end{tabular} & \begin{tabular}[c]{@{}l@{}}48305.003 / 21133.46 \\23.069 / 10723.824\end{tabular} & \begin{tabular}[c]{@{}l@{}}1.374 / 0.246 \\0.032 / 0.008\end{tabular} \\
\hline

\end{tabular}
\end{adjustbox}
\end{table*}
\clearpage

\section{CONCLUSION}
In this paper, we evaluate the robustness of Visual SLAM methods in different environments. It is shown that from the compared methods ORB-SLAM2 is the most robust with respect to the environment. The inability of precise scale prediction is demonstrated for monocular methods. All methods displayed variance to speed, making it an important factor in further research. Also, the effect of SE(3) Umeyama alignment is shown to improve the accuracy with all the explored methods. 

\section{FUTURE WORK}
In the future the authors would like to evaluate the Visual SLAM methods in gravel and off-road scenarios. These environments would provide us with the insights of how the current methods perform in not so ideal conditions. Additionally, it would be interesting to evaluate the methods the capabilities in rain, snow and fog.

\section*{ACKNOWLEDGMENT}
This project and research is supported by Archimedes Foundation and Milrem Robotics under the Framework of Support for Applied Research in Smart Specialization Growth Areas.


{\small
\bibliographystyle{ieee_fullname}
\bibliography{egpaper}
}

\end{document}